%% file: main.tex
\PassOptionsToPackage{table,dvipsnames}{xcolor}

\documentclass[10pt,twocolumn,letterpaper]{article}

\usepackage{cvpr}              

\input{preamble}

\definecolor{cvprblue}{rgb}{0.21,0.49,0.74}
\usepackage[pagebackref,breaklinks,colorlinks,allcolors=cvprblue]{hyperref}

\def\paperID{****} 
\def\confName{CVPR}
\def\confYear{2026}

\title{Measuring the (Un)Faithfulness of Concept-Based Explanations}

\author{Shubham Kumar\\
University of Illinois Urbana-Champaign\\
{\tt\small sk138@illinois.edu}
\and
Narendra Ahuja\\
University of Illinois Urbana-Champaign\\
{\tt\small n-ahuja@illinois.edu}
}

\begin{document}
\maketitle
\input{sec/0_abstract}   
\input{sec/1_intro}
\input{sec/1-5_formalized}
\input{sec/2_relatedworks}
\input{sec/3_method}
\input{sec/4_experiments}
{
    \small
    \bibliographystyle{ieeenat_fullname}
    \bibliography{main}
}

\input{sec/supp}

\end{document}

%% file: preamble.tex
\usepackage{multirow}
\usepackage{adjustbox} 
\usepackage{dsfont} 
\usepackage{pifont} 
\newcommand{\g}{\cellcolor{gray!20}}
\newcommand{\cmark}{\ding{51}}
\newcommand{\xmark}{\ding{55}}

\definecolor{GoodGreen}{HTML}{2E7D32} 
\definecolor{BadRed}{HTML}{C62828}     

\newcommand{\ck}[1]{\textcolor{#1}{\cmark}}
\newcommand{\xx}[1]{\textcolor{#1}{\xmark}}

%% file: sec/0_abstract.tex
\begin{abstract}
Deep vision models perform input-output computations that are hard to interpret. Concept-based explanation methods (CBEMs) increase interpretability by re-expressing parts of the model with human-understandable semantic units, or \textit{concepts}. Checking if the derived explanations are \textit{faithful}---that is, they represent the model's internal computation---requires a \textit{surrogate} that combines concepts to compute the output. Simplifications made for interpretability inevitably reduce faithfulness, resulting in a tradeoff between the two. State-of-the-art unsupervised CBEMs (U-CBEMs) are seemingly more interpretable, while also being more faithful to the model. However, we observe that the reported improvement in faithfulness artificially results from either (1) using overly complex surrogates, which introduces an unmeasured cost to the explanation's interpretability, or (2) relying on deletion-based approaches that, as we demonstrate, do not properly measure faithfulness. We propose \textbf{Surrogate Faithfulness (SURF)}, which (1) replaces prior complex surrogates with a simple, linear surrogate that measures faithfulness without changing the explanation's interpretability and (2) introduces well-motivated metrics that assess loss across all output classes, not just the predicted class. We validate SURF with a measure-over-measure study by proposing a simple sanity check---explanations with random concepts should be less faithful---which prior surrogates fail. SURF enables the first reliable faithfulness benchmark of U-CBEMs, revealing that many visually compelling U-CBEMs are not faithful. \href{https://github.com/skumar-ml/surf-eval}{\textbf{Code is released.}}
\end{abstract}

%% file: sec/1_intro.tex
\section{Introduction}
\label{sec:intro}
Deep learning models have delivered state-of-the-art (SOTA) results across diverse tasks, yet their internal computation remain difficult to interpret \cite{Gerlings2020ReviewingTN}, which is especially problematic in high-stakes domains such as healthcare and finance. This gap has given rise to explainable AI (XAI), whose goal is to produce \textit{explanations} of model behavior. Two properties characterize explanation quality: \textit{interpretability}—explanations that humans can understand—and \textit{faithfulness}—explanations that reflect the model’s internal computation. XAI methods must balance the natural tension between interpretability and faithfulness. While interpretability is assessed with human studies, faithfulness must be evaluated by defining a \textit{surrogate} that maps the explanation to the model’s outputs and measuring the loss between the surrogate and model. Since explanations are inevitably lossy, some discrepancy is expected. For certain XAI methods, defining a surrogate is challenging, so alternative measures, or \textit{proxies}, attempt to evaluate faithfulness using different criteria.

One broad family of XAI methods constructs \textit{inherently interpretable models}, injecting an explanatory structure directly into the model \cite{protopnet, senn}. Merging the explanation and the model eliminates the need for faithfulness checks; however, these methods often underperform black-box models by reducing model complexity (capacity) for explainability. In contrast, \textit{feature attribution} methods express a trained model's outputs in terms of each input feature's contribution, commonly done by finding the output's sensitivity to changes in input features, either through gradients or permutations \cite{wang2024gradient, lime, shapley, Petsiuk2018rise, sobol}. For images, attributions are visualized as pixel-level heatmaps. Although such explanations are intuitive, developing faithfulness measures has been challenging, since many attribution methods do not specify a surrogate to reconstruct the model's output from the explanation. Thus, proxies---such as checking if important pixels come from foreground regions or looking for significant model degradation after deleting important pixels---are used to study faithfulness. Despite strong performance on the proxies, clever sanity checks have revealed faithfulness problems \cite{Adebayo2018SanityCF, Frye2020ShapleyManifold, Sixt2019WhenEL, Hase2021TheOP}. The resulting heatmaps also lack interpretability---they only indicate \emph{where} the model attends, not \emph{what} semantics it recognizes \cite{Nguyen2021TheEO, colin2022what}. 
 
\textit{Concept-based explanation methods} (CBEMs) improve interpretability by explaining predictions in terms of human-understandable \textit{concepts} (e.g., edges, colors, object parts) \cite{net2vec, netdiss, kimTCAV, zhou2014object}. To recognize concepts, supervised CBEMs require a concept-annotated image dataset, which is costly and subjective; moreover, any fixed concept vocabulary may bias the explanation and miss important aspects of the model’s computation, raising faithfulness concerns \cite{Ramaswamy2022OverlookedFI}. To avoid these issues, \textit{unsupervised} CBEMs (U-CBEMs) discover concepts automatically as \textit{concept activation vectors} (CAVs)—directions in the model's intermediate representation space—paired with a \textit{concept importance} score that captures the concept's relevance to the output. Since CBEMs operate on the model's intermediate representations, defining a surrogate is more natural, paving the way for proper faithfulness evaluations.

However, a closer look shows that U-CBEMs have adopted inadequate faithfulness measures. We discover that existing evaluations rely on complex surrogates, which allow U-CBEMs to simultaneously show users interpretable explanations and report high faithfulness, yet the explanations do not clearly lead to the model's output. Furthermore, U-CBEMs adapt popular deletion-based proxies from the attribution literature, yet \cref{sec:related-deletion-measures} outlines serious unresolved limitations that prevent them from properly measuring faithfulness. Finally, \textit{each} work that proposes a new U-CBEM also evaluates it with a \textit{new} faithfulness measure, with no measure-over-measure comparison, indicating a lack of consensus on faithfulness measures in the field.

We argue that these issues have misled us into believing that current U-CBEMs are faithful. To shed light on this, we make the following contributions:

\begin{enumerate}
    \item \textbf{Organize prior U-CBEM faithfulness measures} under a common framework, allowing us to discuss their limitations and identify appropriate desiderata.
    \item \textbf{Propose Surrogate Faithfulness (SURF)}, a faithfulness measure satisfying the desiderata. Our \textit{measure-over-measure comparison} checks if faithfulness decreases as the explanation is increasingly randomized; only SURF passes this check.
    \item \textbf{Conduct the first, comprehensive faithfulness benchmark of current U-CBEMs} across a variety of tasks and model architectures. SURF shows that SOTA U-CBEMs previously evaluated to be faithful are not.
    \item Leverage SURF to \textbf{provide a selection criterion for the number of concepts} U-CBEMs discover, improving on prior work that sets this hyperparameter arbitrarily.
\end{enumerate}

%% file: sec/1-5_formalized.tex
\section{Preliminaries}
We first introduce notation used throughout the paper, as well as preliminaries to familiarize the reader with the field. 

\smallskip

\noindent \textbf{Background:} Model $\phi : \mathcal{X} \to \mathcal{Y}$ maps from an input space $\mathcal{X} \subseteq \mathbb{R}^P$ to an output space $\mathcal{Y} \subseteq \mathbb{R}^C$. Assume $\phi$ admits an intermediate space $\mathcal{H} \subseteq \mathbb{R}^D$. Let $g : \mathcal{X} \to \mathcal{H}$ and $f : \mathcal{H} \to \mathcal{Y}$. Thus, $\mathbf{y} = \phi(\mathbf{X}) = f(g(\mathbf{X}))$. Let $g(\mathbf{x}) = \mathbf{h}  \in \mathcal{H}$ represent the embedding of $\mathbf{x}$ (e.g., image patch), and $g(\mathbf{X}) = \mathbf{H} \in \mathbb{R}^{(HW) \times D}$ denote the vectorized application of $g$ on $\mathbf{X} = \begin{bmatrix} \mathbf{x}_1 & \cdots & \mathbf{x}_{HW} \end{bmatrix}^T$, where $H, W$ are the spatial dimensions.

Let $E$ be an explanation function, and let $E(\mathbf{X}) \in \mathcal{E} $ represent the explanation for an $\mathbf{X} \in \mathcal{X}$ (e.g., image). Let $s : \mathcal{E} \to \mathcal{Y}$ be a surrogate. Motivated by Ribeiro et al. \cite{lime}, $E$ is completely faithful to $\phi$ if $s$ reproduces $\phi(\mathbf{X}) \; \forall \; \mathbf{X} \in \mathcal{X}$. Thus, a faithfulness measure is defined by a choice of surrogate $s(.)$ and metric $d(.)$ as:
\begin{equation}
    Faith(E; d, s) = \int_{\mathcal{X}} d(\phi(\mathbf{X}), s(E(\mathbf{X}))) d\mathbf{X}
\end{equation}
\noindent \textbf{U-CBEMs:} To generate explanations, U-CBEMs find, for all outputs $i \in C$, a set of $K$ CAVs $V_i = \{\mathbf{v}_{i,k}\}_{k=1}^{K}$ and concept importances $A_i = \{\alpha_{i,k}\}_{k=1}^{K}$. Note that while, for a given input, $K$ can vary per output class, all prior works find the same $K$ per output class. U-CBEMs also define a mechanism $\mathcal{P}(\mathbf{h}; V) : \mathcal{H} \to \mathbb{R}^{K}$ to project embedding $\textbf{h}$ to the concept space defined by CAVs in $V$. The explanation for output $i$ is:
\begin{equation}
\begin{aligned}
    E_i(\mathbf{X}) &= E_i(\mathbf{X} ; g, V_i, A_i, \mathcal{P}) \\
    &= \{\mathcal{P}(g(\mathbf{X}); V_i), A_i\}
\end{aligned}
\end{equation}
$E_i(\mathbf{X})$ is then conveyed to the user through some visual interface. Note that $\mathcal{P}(g(\mathbf{X}))$ denotes the vectorized application of $\mathcal{P}$ on $g(\mathbf{X}) = \mathbf{H}$.

Following prior works, we study faithfulness in the final layer, where the spatial dimensions $H=W=1$, so $\mathbf{H} = \mathbf{h}$. Given that U-CBEMs operate on $g(\mathbf{X})$, the faithfulness measure can be simplified to:
\begin{equation}
\begin{aligned}
\label{eq:faithfulness-ucbem-def}
    Faith_{\text{U-CBEM}}(\mathcal{P}, \{V_i\}_{i=1}^C, \{A_i\}_{i=1}^C; d, s) = \\
    \int_{\mathcal{H}} d(f(\mathbf{h}), \{s(\mathcal{P}(\mathbf{h}; V_i), A_i)\}_{i=1}^C) d\mathbf{h}
\end{aligned}
\end{equation}

%% file: sec/2_relatedworks.tex
\begin{figure*}[tb]
    \centering
    \includegraphics[width=0.9\linewidth]{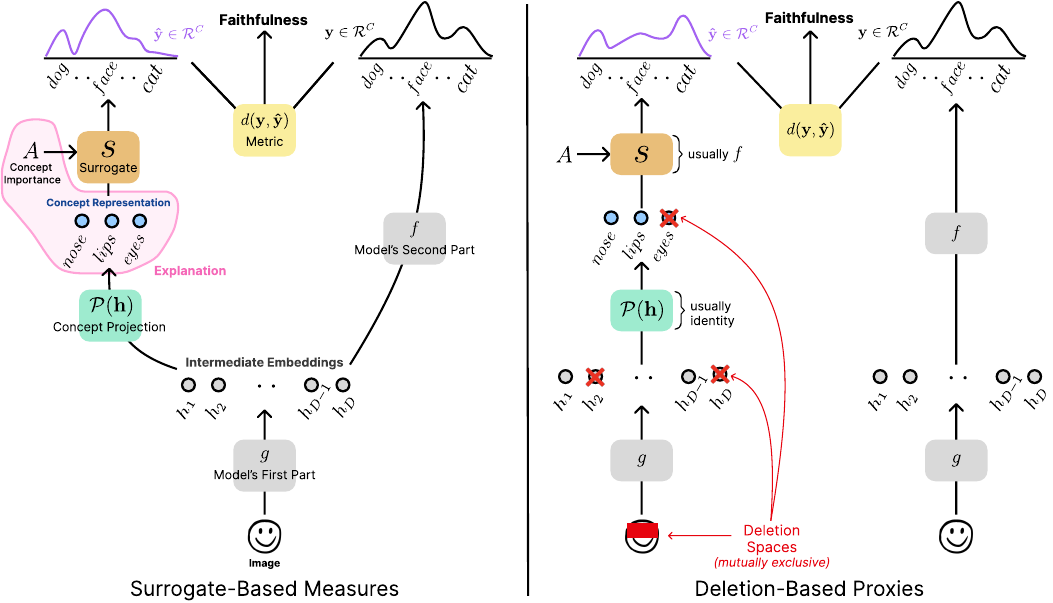}
    \caption{\textbf{Framework.} U-CBEM faithfulness measures compare (using metric $d$) the output $\mathbf{y}$ of the original model $\phi(.) = f(g(.))$ with the output $\mathbf{\hat{y}}$ from the explanation. To obtain the explanation, U-CBEMs transform intermediate representation $\mathbf{h} \in \mathcal{R}^D$ to the concept representation through concept projection $\mathcal{P}(\mathbf{h})$ and provide an accompanying concept importance $A$. The explanation is then passed through surrogate $s$ to obtain $\mathbf{\hat{y}}$. Deletion-based proxies (right) observe model degradation after performing deletion in a deletion space. Surrogate-based measures (left) do not manipulate the computation; instead, they introduce a surrogate to directly approximate \cref{eq:faithfulness-ucbem-def}.}
    \label{fig:framework-overview}
\end{figure*}

\section{Related Works on U-CBEMs}
\label{sec:relatedworks}
Wang et al. \cite{Wang2015UnsupervisedLO} find CAVs by K-Means clustering intermediate embeddings computed on a dataset. Each cluster is associated with a semantic concept (e.g., car tire, window, headlight) by tracing back the clustered embeddings to their corresponding image patches. Ghorbani et al. \cite{Ghorbani2019ACE} improve this with ACE, which finds concepts at different scales by hierarchically segmenting images into ``candidate" segments and clustering the embeddings of each candidate. To ensure that the discovered CAVs fully represent the model's behavior, Yeh et al. \cite{ConceptSHAP} propose ConceptSHAP (C-SHAP), which finds CAVs that maximize a proposed completeness score; they use Shapley Values \cite{Young1985MonotonicSO} to find concept importance. ICE \cite{Zhang2020InvertibleCE} extends ACE by replacing clustering with Non-negative Matrix Factorization (NMF), resulting in a parts-based explanation. The authors show that concepts found with NMF are more interpretable than those found by K-Means or Principal Component Analysis (PCA). Building on this, Fel et al. \cite{Fel2022CRAFTCR} present CRAFT, which recursively applies NMF through the model to decompose concepts into sub-concepts, and they use sensitivity analysis for finding concept importance. CDISCO \cite{graziani2023concept} uses the singular value decomposition to find CAVs and a gradient-based method for finding concept importance. MCD \cite{vielhaben2023multidimensional} extends the notion of CAVs to a multidimensional linear subspace, rather than a single vector, allowing it to describe more of the model's behavior with fewer concepts. Finding issues with MCD's interpretability, HU-MCD \cite{hu-mcd} incorporates Segment Anything Model \cite{SAM} to discover more interpretable concepts. Recently, sparse autoencoders (SAEs) have emerged as a promising, scalable U-CBEM for large language models, and recent attempts have applied them to interpret vision models \cite{gorton2024the, lim2025sparse, fel2025archetypal}.

CBEMs and U-CBEMs should not be confused with Concept Bottleneck Models (CBMs), which create an inherently explainable concept model meant to replace the original black-box model \cite{pmlr-cbm-koh20a, prob-cbm, posthoc-cbm, labelfree-cbm}.

\begin{table*}[t]
\caption{\textbf{Prior deletion-based proxies (top) \& surrogate-based measures (bottom)} are organized according to their differing factors: \textit{Deletion Space} (DS), \textit{Deletion Method} (DM), \textit{Metric}, \textit{Surrogate}, and \textit{Concept Projection} (CP). Each prior work uses a different proxy or measure, with no measure-over-measure comparison.} 
\label{tab:deletion-table}
\small
\centering
    \begingroup
        \footnotesize
        \begin{tabular}{lccccc}
            \toprule
            \textbf{Method} & \textbf{DS} & \textbf{DM} & \textbf{Metric} & \textbf{Surrogate} & \textbf{CP} \\
            \midrule
            ACE & Pixel & Constant (grey) & Class Accuracy & Original Model & Identity \\
            MCD & Pixel & Inpainting & Class Accuracy & Original Model & Identity \\
            HU-MCD & Pixel & Masking & Class Accuracy & Original Model & Identity \\
            CDISCO (M1) & Pixel & Constant (grey) & \# Classes with $>80\%$ Accuracy Loss & Original Model & Identity \\
            CDISCO (M2) & Weight & Constant (zero) & Class Accuracy & Original Model & Identity \\
            CRAFT & Concept & Constant (zero) & Class Logit & Reconstruct $\to$ Original Model & U-CBEM \\            
            \midrule
            \multirow[c]{2}{*}{ICE} & \multirow[c]{2}{*}{--} & \multirow[c]{2}{*}{--}
              & Normalized Target Class L1 Logit Loss & \multirow[c]{2}{*}{U-CBEM Reconstruct $\to$ Original Model} & \multirow[c]{2}{*}{U-CBEM} \\
             &  &  & Top-1 Accuracy & \\[2pt]
            C-SHAP & -- & -- & Normalized Top-1 Accuracy & MLP Reconstruct $\to$ Original Model & U-CBEM \\
            \bottomrule
        \end{tabular}
    \endgroup
\end{table*}


\section{Unifying U-CBEM Faithfulness Measures}
\label{sec:related-metrics}
Each U-CBEM mentioned in \cref{sec:relatedworks} introduces a \textit{different faithfulness measure}, indicating a lack of consensus and preventing fair comparisons of faithfulness across U-CBEMs. We unify faithfulness measures under a common framework, shown in \cref{fig:framework-overview}. Any evaluation measure compares the output of the original model with the output from the explanation (\textcolor{blue}{\textit{Metric}}). To obtain the output from the explanation, a \textcolor{blue}{\textit{Concept Projection}} transforms the intermediate representation to a concept representation, which is then passed through a \textcolor{blue}{\textit{Surrogate}}. Deletion-based proxies assert that removing the most important concepts will result in the greatest model degradation; more degradation is interpreted as a signal for higher faithfulness. Alternatively, surrogate-based measures avoid changing the input and introduce a surrogate to directly approximate faithfulness as in \cref{eq:faithfulness-ucbem-def}. Appendix \cref{sec:appendix-sae-discussion} offers a discussion and comparison with faithfulness metrics introduced for SAEs.

\subsection{Deletion-Based Proxies}
\label{sec:related-deletion-measures}
To perform deletion, deletion-based proxies must remove concepts \textit{(\textcolor{blue}{Deletion Method})} from the input's \textit{\textcolor{blue}{Deletion Space}} in decreasing order of importance.

\textit{Deletion Space:} Where concepts are removed. This is either the image space, the model's weight space, or the concept space (shown in \cref{fig:framework-overview} (right)).

\textit{Deletion Method:} How concepts are removed from the \textit{Deletion Space}, usually by setting the concept to a baseline value (e.g., 0).

Except for CRAFT, proxies set the concept projection to identity and the surrogate to $f(.)$ from the original model. Prior proxies (organized in \cref{tab:deletion-table}) make different choices along each \textit{\textcolor{blue}{italicized factor}} mentioned above. \textbf{ACE} deletes concepts in the image space by setting associated pixels to a constant value (Constant Deletion). They measure the model's classification accuracy after each deletion across a dataset. \textbf{MCD} measures classification accuracy after deleting concepts in the pixel space with inpainting. \textbf{HU-MCD} similarly measures classification accuracy but performs deletions through a masking strategy which simulates running the model on the irregularly-shaped input, better ignoring masked regions. \textbf{CDISCO} has two measures. The first measure (M1) makes Constant Deletions on pixels and measures the number of classes whose accuracy degrades by more than 80\%. The second measure (M2) deletes concepts by zeroing out model parameters associated with the concept, and they report classification accuracy. \textbf{CRAFT} uses the U-CBEMs defined $\mathcal{P}$ to transform $\mathbf{h}$ to a concept representation. It makes Constant Deletions in the concept space by setting the concept to zero; then, it reconstructs $\mathbf{h}$ from the perturbed concept representation and measures the change in the corresponding class's logit score. 

There are two notable issues with deletion-based proxies that prevent them from accurately evaluating faithfulness. \textbf{First, it is unclear how to delete a concept.} Concepts are usually deleted by Constant Deletion, commonly with a baseline of \textbf{0}. This baseline must entirely delete the presence of the concept without affecting other concepts. However, the feature attribution literature has shown that common baseline choices do not guarantee that a feature is completely deleted without affecting other features, leading to unfaithful explanations \cite{Frye2020ShapleyManifold}. Removing features by marginalization is more accurate but requires evaluating expensive, high-dimensional expectations.

\textbf{Second, there are no guarantees that representations in the Deletion Space will stay on manifold after deletion.} The feature attribution literature \cite{Frye2020ShapleyManifold, akata2022manifold, taufiq2023manifold, Slack2019FoolingLA, yeh2022threading} has found that the perturbed inputs used for creating explanations may be unrealistic, lying off the data manifold. In the off-manifold regions of the data space, a highly non-linear model behaves unpredictably, so the model's outputs may not be meaningfully related to input features. Thus, calculating feature importance with off-manifold inputs can lead to incorrect explanations \cite{Frye2020ShapleyManifold}. More seriously, these explanations can be adversarially attacked; in \cite{Slack2019FoolingLA}, a model's off-manifold behavior was modified to hide its dependence on undesirable features used for on-manifold inputs. Such issues can be mitigated by using on-manifold input perturbations, but this requires accurate modeling of the input distribution \cite{Frye2020ShapleyManifold, taufiq2023manifold, vu2022emap, zaher2024manifold, Hase2021TheOP}. These problems are not unique to feature attribution; they extend to deletions in the concept and weight space, which make similar manipulations in the model's intermediate representation space. 

Thus, we argue that deletion-based proxies are unreliable, and we avoid using or benchmarking against them.

\subsection{Surrogate-Based Measures}
\label{sec:surrogate-related-works}
Any surrogate-based measure must choose the surrogate $s$ and the metric $d$. For these measures, the input image, and thus the intermediate representation, is the same in both paths of \cref{fig:framework-overview} (left), and the concept projection is defined by the U-CBEM being evaluated.

\subsubsection{Desiderata}
The faithfulness definition in \cref{eq:faithfulness-ucbem-def} does not constrain the surrogate $s$, but there are certain desiderata our surrogate should satisfy. \textbf{First}, $s$ should be as simple as possible. Recognize that $s$ reflects the mental computation any human interpreter is expected to do when trying to connect the explanation to the model's prediction. An interpretable explanation that requires a complex $s$ just shifts the interpretation issue downstream; the human interpreter still lacks insight into how the model reaches its prediction. \textbf{Second}, $s$ should incorporate all components of the explanation. For example, U-CBEMs explain with both $V_i$ and $A_i$, so both should be used by $s$. If a component is not included, its impact on faithfulness cannot be measured. \textbf{Third}, we desire metric(s) $d$ that rewards explanations that closely reconstruct \textit{all} of $\mathbf{y}$ (e.g. across all classes), not just specific (e.g., predicted) classes.

\subsubsection{Prior Surrogate-Based Measures}
Two prior works have proposed surrogate-based measures: 

\smallskip

\textbf{ICE-Eval:} ICE introduces its own evaluation measure (we term as ICE-Eval). ICE-Eval's surrogate assumes the U-CBEM's projection operation $P$ has a reconstruction function $\tilde{\mathcal{P}}^{-1} : \mathbb{R}^{K} \to \mathcal{H}$. Their surrogate then is:
\begin{equation}
    \begin{aligned}
            \mathbf{\hat{y}}_i = s_i(.) &= f_i(\tilde{\mathcal{P}}^{-1}(\mathcal{P}(\mathbf{h}; V_i))) \;\; \forall \; i             
    \end{aligned}
\end{equation}
where $f_i(.)$ denotes the $i$'th output from function $f(.)$. ICE-Eval uses two metrics (only applicable to classification):
\begin{equation}
\label{eq:ICE-norm-logit-error}
    ICE_1 = \frac{1}{|\mathcal{V}|} \sum_{\mathbf{x},t \in \mathcal{V}} \frac{|y_t - \hat{y}_t|}{|y_t|}
\end{equation}
where $t$ is the index corresponding to $\mathbf{x}$'s groundtruth class and $\mathcal{V}$ is the set of inputs used to evaluate faithfulness. Letting $\mathbf{p}$ denote class probability scores, the second metric is:
\begin{equation}
\label{eq:ICE-top-1-error}
    ICE_2 = \frac{1}{|\mathcal{V}|} \sum_{\mathbf{x} \in \mathcal{V}} \mathds{1}\{argmax(\mathbf{p}) = argmax(\hat{\mathbf{p}})\}
\end{equation}

\textbf{C-SHAP-Eval:} C-SHAP introduces its own evaluation measure (we term as C-SHAP-Eval). Its surrogate learns a two-layer perceptron $\psi_{\text{C-SHAP}} : \mathbb{R}^{K} \to \mathcal{H}$:
\begin{equation}
    \begin{aligned}
            \mathbf{\hat{y}}_i = s_i(.) &= f_i(\psi_{\text{C-SHAP}}(\mathcal{P}(\mathbf{h}; V_i))) \;\; \forall \; i             
    \end{aligned}
\end{equation}
The metric they define is:
\begin{equation}
\label{eq:CSHAP-top-1-error}
    CSHAP_1 = \frac{\sum_{\mathbf{x},t \in \mathcal{V}} \mathds{1}\{t = argmax(\hat{\mathbf{p}})\}-a_r} 
    {\sum_{\mathbf{x},t \in \mathcal{V}} \mathds{1}\{t = argmax(\mathbf{p})\}-a_r}
\end{equation}
where $a_r$ is the accuracy from random predictions.

ICE-Eval and C-SHAP-Eval fail all three of our desiderata. (1) A critical piece of both surrogates is to \textit{reconstruct} the model's original embedding, and this reconstruction may be non-linear. Additionally, C-SHAP introduces non-linearity and learnable parameters via $\psi_{\text{C-SHAP}}$. The overly complex surrogates place a large burden on the human interpreter, which is not reflected in prior interpretability and faithfulness evaluations. (2) Neither surrogate depends on concept importances ($A_i's$). Thus, unfaithful $A_i$'s will \textit{not} alter faithfulness scores. (3) Error is measured only on the predicted or groundtruth class, ignoring errors on the remainder of the output distribution.

%% file: sec/3_method.tex
\begin{table}[t]
\caption{\textbf{Surrogate-based desiderata} We compare prior surrogate-based measures introduced in ICE and C-SHAP to the proposed Surrogate Faithfulness (SURF) measure. According to our desiderata, we require surrogates that are simple (e.g., not reconstruction-based and not requiring additional parameters), incorporate concept importances into the surrogate, and measure errors across all outputs. Prior measures largely fail to meet these desiderata. We mark in \textbf{\textcolor{GoodGreen}{green}} and \textbf{\textcolor{BadRed}{red}} the properties that meet our desiderata. A \cmark \enspace or \xmark \enspace denotes that the method does or does not have the property.}
\label{tab:surrogate-differences}
\small
  \centering
  \begin{tabular}{lccc}
    \toprule
    \textbf{Property} & \textbf{ICE} & \textbf{C-SHAP} & \textbf{SURF (Ours)} \\
    \midrule
    Reconstruction-Based? & \ck{BadRed} & \ck{BadRed} & \xx{GoodGreen} \\
    Additional Parameters? & \xx{GoodGreen} & \ck{BadRed} & \xx{GoodGreen} \\
    Uses Concept Importance? & \xx{BadRed} & \xx{BadRed} & \ck{GoodGreen} \\
    Errors on All Outputs? & \xx{BadRed} & \xx{BadRed} & \ck{GoodGreen}\\
    \bottomrule
  \end{tabular}
\end{table}

\begin{table*}[t]
    \centering
    \caption{\textbf{Measure-over-measure comparison.} We compare surrogate-based measures across three explanation settings (\textit{Perfect}, \textit{Rand Imp}, and \textit{Full Rand}). We expect evaluations in the \textit{Perfect} setting to give perfect faithfulness scores, and we expect progressively worse faithfulness evaluations as we increase randomness (just importances in \textit{Rand Imp} and fully random in \textit{Full Rand}). We include $\text{SURF}_{\text{EMD}}$ and $\text{SURF}_{\text{MAE}}$ along with metrics used by prior works. The SURF metrics behaves as expected, whereas C-SHAP-Eval reports higher faithfulness in the \textit{Full Rand} setting and ICE-Eval reports perfect faithfulness scores in the \textit{Rand Imp} setting.}
    \small
    \begingroup
        \footnotesize
        \begin{tabular}{c@{\hskip 6pt}lccccccc}
            \toprule
            & \textbf{Surrogate} & \textbf{Top-1} (\%) ($\uparrow$) & \textbf{Rank Corr} ($\uparrow$) & \textbf{Norm L1} ($\downarrow$) & $\text{\textbf{SURF}}_{\text{MAE}}$ ($\downarrow$) & $\text{\textbf{SURF}}_{\text{EMD}}$ ($\downarrow$) & \textbf{Params Learnt} ($\downarrow$) & \textbf{FLOPs} ($\downarrow$) \\
            \midrule
            
            \multirow{4}{*}{\rotatebox{90}{\underline{Perfect}}}
            & C-SHAP-Eval (CEL) & 9.02 & -0.02 & 1.27 & 1.97 & 0.865 & 1M & 205M \\
            & C-SHAP-Eval (L1) & 6.13 & 0.08 & 2.15 & 0.54 & 0.883 & 1M & 205M \\
            & ICE-Eval & 100 & 1.00 & 0.00 & 0.00 & 0.000 & 0 & 614K\\
            & \g SURF (Ours) & \g 100 & \g 1.00 & \g 0.00 & \g 0.00 & \g 0.000 & \g 0 & \g 200\\
    
            \midrule
    
            \multirow{4}{*}{\rotatebox{90}{\underline{Rand Imp}}}
            & C-SHAP-Eval (CEL) & 9.02 & -0.02 & 1.27 & 1.97 & 0.865 & 1M & 205M \\
            & C-SHAP-Eval (L1) & 6.13 & 0.08 & 2.15 & 0.54 & 0.883 & 1M & 205M \\
            & ICE-Eval & 100 & 1.00 & 0.00 & 0.00 & 0.000 & 0 & 614K \\        
            & \g SURF (Ours) & \g 97.5 & \g 0.13 & \g 0.83 & \g 2.70 & \g 0.862 & \g 0 & \g 200 \\
    
            \midrule
    
            \multirow{4}{*}{\rotatebox{90}{\underline{Full Rand}}}
            & C-SHAP-Eval (CEL) & 97.6 & 0.02 & 181.7 & 168.2 & 0.125 & 1M & 205M \\
            & C-SHAP-Eval (L1) & 6.1 & 0.08 & 3.59 & 1.721 & 0.883 & 1M & 205M \\
            & ICE-Eval & 3.3 & 0.00 & 1.00 & 3.17 & 0.882 & 0 & 614K \\
            & \g SURF (Ours) & \g 1.3 & \g 0.00 & \g 1.01 & \g 3.17 & \g 0.883 & \g 0 & \g 200 \\
            \bottomrule        
        \end{tabular}
    \endgroup
    \label{tab:justify-surrogate}
\end{table*}

\section{Method: Surrogate Faithfulness (SURF)}
\label{sec:method}
Surrogate Faithfulness (SURF) introduces a simple, linear surrogate and two metrics; taken together, they allow SURF to satisfy the desiderata.
\subsection{The SURF Surrogate}
\label{sec:method-surrogate}
The final linear layer $\mathbf{F} = \begin{bmatrix} \mathbf{F}_1 & \cdots & \mathbf{F}_C \end{bmatrix}$ operates on representation $\mathbf{H}$ to give the output $\mathbf{y} \in \mathcal{Y}$. In terms of their components, $\mathbf{H} = \begin{bmatrix} \mathbf{h}_1 & \cdots & \mathbf{h}_{HW} \end{bmatrix}^T$ and $\mathbf{F}_i = \begin{bmatrix} \mathbf{f}_{i, 1} & \cdots & \mathbf{f}_{i, HW} \end{bmatrix}^T \in \mathbb{R}^{(HW) \times D}$. Thus, output $y_i$ of $\phi$ is given by (up to the bias term): 
\begin{equation}
\begin{aligned}
    y_i &= \sum_{j=1}^{HW} \mathbf{h}_j^T \mathbf{f}_{i, j} =  \sum_{j=1}^{HW} \mathbf{h_j}^T \frac{\mathbf{f}_{i, j}}{{\|\mathbf{f}_{i, j}\|}_2} {\|\mathbf{f}_{i, j}\|}_2 \\
    &\triangleq \sum_{j=1}^{HW} \mathbf{h}_j^T \mathbf{v}_{i, j} \alpha_{i,j} \quad \text{where } {\|\mathbf{v}_{i,j}\|}_2 = 1 \quad \forall\ i,j
\label{eq:linear-generalized}
\end{aligned}
\end{equation}
where $\alpha_{i,j} = {\|\mathbf{f}_{i,j}\|}_2$. In words, the model's linear layer projects each embedding $\mathbf{h}_j$ onto a learned, class-specific direction (or \textit{CAV}) $\mathbf{f}_{i,j}$. The projection is scaled by the norm (or \textit{importance}) $\alpha_{i,j}$ of the learned direction. If the model uses global pooling to reduce the final embedding of $\mathbf{x}$ to a single vector, the summation over $j$ above can be omitted.

We define SURF's surrogate $s$ to take the form of (\cref{eq:linear-generalized}), replacing $\mathbf{F}$ with the concept representation and $A_i$'s obtained from any U-CBEM. The surrogate is:
\begin{equation}
\label{eq:sf-add}
    \begin{aligned}
            \mathbf{\hat{y}}_i &= s_i(.) = \sum_{j=1}^{HW} \sum_{k=1}^{K} \alpha_{i,k} \mathcal{P}(\mathbf{h}_{j}; V_i)_k \quad \forall\ i             
    \end{aligned}
\end{equation}
where $\mathcal{P}(.)_k$ denotes the $k$'th element. Following prior U-CBEMs and evaluation measures, SURF also operates only on the model's final linear layer. Thus, its surrogate fully represents the model's computation, while not introducing any trainable parameters and greatly reducing complexity compared to prior faithfulness surrogates.

\subsection{The SURF Metrics}
The SURF metrics measure the difference in model and surrogate outputs. For classification models, this means the logits. However, a model's logit-space is unconstrained, varying drastically between models. This makes logit-space metrics difficult to interpret. Furthermore, metrics in the logit-space dilute the importance of the predicted class by aggregating errors over all classes. On the other hand, the probability-space (after softmax) normalizes the logits, allowing for a bounded, interpretable metric. However, low error in the probability-space \textit{does not} always imply low error in the logit-space. Due to the softmax, the probability-space emphasizes the predicted class and diminishes the other classes, so one could achieve a low error by accurately reproducing the predicted class and ignoring the others. Since each space addresses the other's flaws, SURF measures errors in both spaces.

If $\phi$ is a classification model, then $\mathbf{y}$ denotes class logits. \textbf{To measure errors in the logit-space}, we use the mean absolute error between the logits:
\begin{equation}
\label{eq:logits-metric}
    \text{SURF}_{\text{MAE}} = \frac{1}{|\mathcal{V}|C} \sum_{\mathbf{x} \in \mathcal{V}}  \sum_{i=1}^{C} |y_i - \hat{y}_i|
\end{equation}
\textbf{To measure errors in the probability-space}, we find the Earth Mover's Distance using a constant distance cost.
\begin{equation}
\label{eq:sf-metric}
    \text{SURF}_{\text{EMD}} = \frac{1}{2|\mathcal{V}|} \sum_{\mathbf{x} \in \mathcal{V}}  \sum_{i=1}^{C} |p_i - \hat{p}_i|
\end{equation}

We choose these specific metrics as they are easily interpretable. A faithful U-CBEM will have explanations with low $\text{SURF}_{\text{EMD}}$ \textit{and} $\text{SURF}_{\text{MAE}}$ errors. For regression, we only use $\text{SURF}_{\text{MAE}}$. 

In contrast to prior surrogate-based measures, SURF meets all of our desiderata. (1) It introduces a simple surrogate with no additional learnable parameters: the output for any class $i$ is a linear combination of the concept representation, weighted by the concept importances ($A_i$). The surrogate \textit{is not reconstruction-based}; it tries to linearly predict the \textbf{next} representation (e.g., class logits), instead of reconstructing embedding $\mathbf{h}$. Thus, only concepts useful for the prediction are needed, which may be a subset of the concepts needed for reconstructing $\mathbf{h}$ from $\mathcal{P}(\mathbf{h}; V)$. (2) SURF's surrogate incorporates $A_i$, so inaccuracies in $A_i$ impact faithfulness scores. (3) SURF defines two metrics $d$ that measure errors across the entire output space, instead of specific classes, to comprehensively assess faithfulness. These differences are summarized in \cref{tab:surrogate-differences}. 

%% file: sec/4_experiments.tex
\begin{table*}[tb]
  \caption{\textbf{Benchmarking U-CBEM faithfulness.} We apply SURF to evaluate explanations from prior U-CBEMs in three tasks. Along with the $\text{SURF}_{\text{MAE}}$ and $\text{SURF}_{\text{EMD}}$, we report other metrics to serve as a comparison. We find that prior U-CBEMs are not faithful, as indicated by large errors in the logit and probability space.}
  \label{tab:benchmark}
  \centering
  \small
  \begingroup  
  \footnotesize

  \begin{subtable}[t]{0.62\textwidth}
    \centering
    \subcaption{Object classification \\ (ResNet-50)}
    \label{tab:object_clf}
    \begin{tabular}{lccccc}
      \toprule
      \textbf{U-CBEM} & $\text{\textbf{SURF}}_{\text{MAE}}$ ($\downarrow$) & $\text{\textbf{SURF}}_{\text{EMD}}$ ($\downarrow$) &
        \textbf{Top-1 (\%)} ($\uparrow$) & \textbf{Rank Corr} ($\uparrow$) \\
      \midrule      
      CDISCO & 3.40 & 0.932 & 0.2 & 0.002 \\
      ICE & 3.33 & 0.628 & 98.9 & 0.093 \\
      CRAFT & 3.19 & 0.878 & 90.6 & 0.068 \\
      C-SHAP & 3.28 & 0.882 & 6.3 & 0.005 \\
      MCD & 2.60 & 0.426 & 99.4 & \textbf{0.145} \\
      HU-MCD & \underline{1.97} & \underline{0.384} & \textbf{99.7} & \underline{0.149} \\
      SAE & \textbf{1.04} & \textbf{0.195} & 99.2 & 0.366 \\
      \bottomrule
    \end{tabular}
  \end{subtable}%
  \hfill
  \begin{subtable}[t]{0.25\textwidth}
    \centering
    \captionsetup{justification=centering} 
    \subcaption{Multi-attribute prediction \\ (MobileNetV2)}
    \label{tab:celeba}
    \begin{tabular}{cc}
      \toprule
      $\text{\textbf{SURF}}_{\text{MAE}}$ & \textbf{Attr-Acc} (\%) ($\uparrow$) \\
      \midrule
      6.77 & 50.7\\
      5.55 & 76.1 \\
      6.87 & 19.6 \\
      7.75 & 51.0 \\
      \textbf{2.83} & \textbf{96.6} \\
      -- & -- \\
      \underline{3.16} & \underline{81.9} \\
      \bottomrule
    \end{tabular}
  \end{subtable}%
  \hfill
  \begin{subtable}[t]{0.10\textwidth}
    \centering
    \subcaption{Age \\ (ViT)}
    \label{tab:age}
    \begin{tabular}{c}
      \toprule
      $\text{\textbf{SURF}}_{\text{MAE}}$\\
      \midrule
      32.6 \\
      -- \\
      -- \\
      -- \\
      -- \\
      -- \\
      \textbf{3.67} \\
      \bottomrule
    \end{tabular}
  \end{subtable}

\endgroup
\end{table*}

\section{Experiments}
\label{sec:results}
We first perform a measure-over-measure study with a basic sanity check: do faithfulness scores decrease as we progressively randomize the explanation? Only SURF passes this check. Then, we benchmark prior U-CBEMs across three tasks, revealing that SOTA methods are not faithful. Finally, we leverage SURF to analyze the tradeoff between the number of concepts in an explanation and faithfulness.

\subsection{Measure-over-Measure Comparison}
\label{sec:expeirment-justifying}
The fundamental challenge in evaluating faithfulness measures is that there is no ``groundtruth" of faithfulness to compare evaluation results to. To address this, we propose a simple sanity check, which introduces three manufactured settings where the relative faithfulness of explanations is known a priori. Then, we check if the faithfulness measure preserves the relative ordering of scores across settings.

Concretely, we look at faithfulness results for a \textit{perfect} explanation and two \textit{randomized} explanations. To generate the \textit{Perfect} explanation, observe that the most faithful explanation of the model is the model itself. Thus, we set $\{A_i, V_i\}_{i=1}^{C}$ according to the weights of the linear classification layer. In our second setting (termed \textit{Rand Imp}), we keep the perfect CAVs fixed and randomly sample the concept importances. Our final setting (termed \textit{Full Rand}) additionally randomly samples the CAVs. In the \textit{Perfect} setting, we expect to obtain no faithfulness error, regardless of the evaluation measure. Then, we expect to obtain increasingly less faithful scores as we progressively increase randomness (from \textit{Rand Imp} to \textit{Full Rand}).

We apply this sanity check on SURF, ICE-Eval, and C-SHAP Eval for an ImageNet-pretrained ResNet-50 \cite{resnet} finetuned on the Caltech-101 dataset \cite{caltech-101}. Along with the SURF metrics, we report Top-1 Accuracy (\textbf{Top-1}) defined in \cref{eq:ICE-top-1-error} and normalized L1 logit error (\textbf{Norm L1}) defined in \cref{eq:ICE-norm-logit-error}. As a complementary metric to $\text{SURF}_{\text{EMD}}$, we measure Spearman's rank correlation (\textbf{Rank Corr}) between surrogate and model outputs. Rank Corr ignores magnitude differences but penalizes fine-grained changes that result in a different prediction ordering. Finally, we measure surrogate complexity by its learnable parameters and FLOPs from concept representation to output. Since C-SHAP-Eval's surrogate was originally trained with cross entropy loss (CEL), we also train it with L1 loss, as this may align better with the SURF metrics. More details are in Appendix \cref{sec:justify-surrogate-appendix}.

Results are reported in \cref{tab:justify-surrogate}, averaged over 10 seeds. In the \textit{Perfect} setting, both ICE-Eval and SURF report perfect faithfulness, as expected. However, both C-SHAP-Eval variants do not achieve perfect faithfulness across any metric, showing C-SHAP-Eval's limitations. In the \textit{Rand Imp} setting, we observe that C-SHAP-Eval and ICE-Eval do not report differences when compared to the \textit{Perfect} setting; because their surrogates do not use concept importances, using random concept importances do not change results from the \textit{Perfect} setting. Contrast this with SURF, which clearly finds a deterioration in faithfulness across \textit{all} metrics. Finally, in the \textit{Full Rand} setting, C-SHAP-Eval unintuitively reports an improvement in faithfulness compared to the \textit{Perfect} setting. ICE-Eval and SURF report similarly poor faithfulness scores for this setting, as expected.

Next, we examine the metrics. SURF's results in the \textit{Rand Imp} setting highlight a failure case for Top-1. The Top-1 score is strong; however, the other metrics (especially $\text{SURF}_{\text{EMD}}$) reveal that the surrogate output has severe errors on the remainder of the output distribution. Norm L1 also comes with issues; notice how C-SHAP-Eval (CEL) reports a lower error compared to C-SHAP-Eval (L1) in the \textit{Perfect} setting, despite C-SHAP-Eval (L1) directly minimizing the logit L1 error. We attribute this inconsistency to Norm L1's emphasis on \textit{only the groundtruth class}; while the L1 error across \textit{all} classes (as measured by $\text{SURF}_{\text{MAE}}$) has decreased, the normalized error for the \textit{groundtruth} class increased, giving an incomplete picture of faithfulness. 

Of the faithfulness measures, only SURF passes the sanity check while having the lowest surrogate complexity (fewest FLOPs and no learnable parameters). The remainder of the paper uses SURF to study prior U-CBEMs. 

\subsection{Benchmarking U-CBEMs}
\label{sec:results-benchmark}

We evaluate seven prior U-CBEMs (CDISCO, ICE, CRAFT, C-SHAP, MCD, HU-MCD, SAE) on a varied range of tasks and models to demonstrate the applicability of SURF. Specifically, our tasks are: 1) \textit{Object Classification}, 2) \textit{Multi-Attribute Prediction}, and 3) \textit{Age Regression}. In all tasks, U-CBEMs discover 5 CAVs (or subspaces for MCD and HU-MCD) per output. CAVs and importances are found on the training set; the resulting explanations are evaluated on the test set. Dataset, finetuning, U-CBEM implementation details, and good-faith efforts made to adapt U-CBEMs to new tasks are in Appendix \cref{sec:benchmarking-appendix}.

We evaluate on three tasks. \textbf{(1) Object Classification:} We use the Caltech-101-finetuned ResNet-50 and report $\text{SURF}_{\text{MAE}}$, $\text{SURF}_{\text{EMD}}$, Top-1, and Rank Corr. HU-MCD requires architectural modifications, so we can only evaluate it on this task, which uses a compatible ResNet-50. Results on a VGG and InceptionV3 are in Appendix \cref{sec:appendix-additional-benchmark-results}. \textbf{(2) Multi-Attribute Prediction:} Given an image, we finetune a MobileNetV2 \cite{Sandler2018MobileNetV2IR} to predict the presence/absence of attributes on the CelebA dataset \cite{celeba}. We report $\text{SURF}_{\text{MAE}}$ and attribute prediction accuracy (Attr-Acc), which is introduced as a task-specific replacement for Top-1. \textbf{(3) Age Estimation:} Given a human face, we finetune a ViT \cite{VIT} to predict their age (in years) on the UTK-Face dataset \cite{utkface}. Because ViTs have negative activations, U-CBEMs with non-negativity assumptions (i.e., ICE and CRAFT) cannot be used. C-SHAP is omitted as it is only applicable to classification, and MCD is incompatible with ViTs. 

Benchmark results are reported in \cref{tab:benchmark}. In the evaluated settings, we observe that \textbf{no prior U-CBEM is faithful to the original model}. In the \textit{Object Classification} task, the most faithful U-CBEM is SAE with an $\text{SURF}_{\text{EMD}}$ of 0.195, denoting significant errors in the probability-space. Only half of the tested U-CBEMs (ICE, MCD, and SAE) perform significantly above random chance in the \textit{Multi-Attribute Prediction} task. SAE exhibits an average error of 3.67 years in the \textit{Age Estimation} task. Note that this evaluation \textit{solely} focuses on faithfulness and does not make any judgment on the interpretability side of the tradeoff.

We believe that U-CBEMs are unfaithful for two reasons: (1) Other than C-SHAP and SAE, all methods discover class-specific CAVs to reconstruct the embedding. Crucially, the class-specific CAVs are found \textit{only} on class-specific images, so using class $c_1$'s CAVs to reconstruct an image embedding of class $c_2$ will have large reconstruction error (e.g., the U-CBEM is operating out-of-distribution), and therefore, large SURF errors. This issue is exacerbated for CRAFT and ICE, which non-linearly project to the concept space. Though MCD and HU-MCD also find class-specific concepts, they achieve superior faithfulness by using concept subspaces; thus they faithfully explain more of the model with the same number of concepts. (2) Most methods do not satisfy a logit-based completeness criterion, preventing a targeted reproduction of model outputs. C-SHAP, instead, satisfies an accuracy-based completeness criterion, and ICE, MCD, and HU-MCD satisfy the logit-based criterion by including uninterpretable residuals, which cannot be shown in the user explanation.

\begin{figure}[tb]
    \centering
    \includegraphics[width=\linewidth]{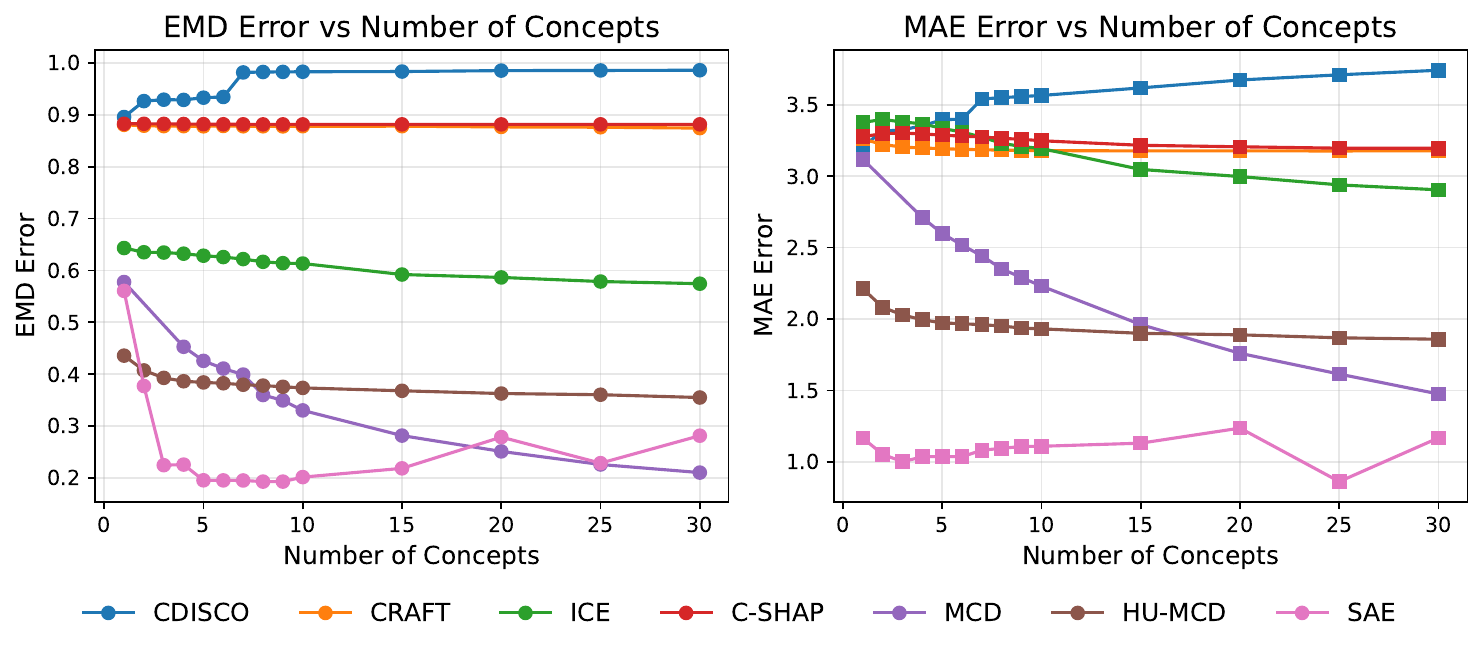}
    \caption{We fit U-CBEMs on \cref{sec:results-benchmark}'s \textit{Object Classification} task with an increasing number of concepts and evaluate their faithfulness with SURF. Some U-CBEMs do not improve as the number of concepts increase. U-CBEMs that improve quickly plateau.}
    \label{fig:faithfulness-parsimony}
\end{figure}

\subsection{Faithfulness vs. Parsimony}
The \textit{most important hyperparameter} for any U-CBEM is the number of concepts to be discovered. Intuitively, having more concepts should result in a more complete (i.e., faithful) explanation. However, explanations containing too many concepts may be difficult to interpret. Cognitive psychology studies show that humans can hold a limited number of items in their working memory at a time \cite{2010TheMM, Miller1956TheMN}. Thus, simple, or \textit{parsimonious}, explanations are preferred, which communicate the bulk of the model's computation with a few concepts. However, inadequate faithfulness measures make the faithfulness-parsimony tradeoff hard to analyze, so prior U-CBEMs commonly set this hyperparameter arbitrarily (10 in ICE, 25 in CRAFT \& ACE). SURF allows us to intelligently analyze this tradeoff.

Using the \textit{Object Classification} task (\cref{sec:results-benchmark}), we fit U-CBEMs with an increasing number of concepts, each evaluated with the SURF measure and visualized in \cref{fig:faithfulness-parsimony}. Some U-CBEMs (CDISCO, CRAFT, C-SHAP, ICE) either marginally improve or perform worse on one or both metrics as they discover more concepts. SAEs initially improve but then oscillates. Only MCD and HU-MCD uniformly improve as the number of concepts increases, and they exhibit a plateauing effect, marking a natural choice for the number of concepts. Interestingly, HU-MCD is more faithful than MCD only when both discover few concepts; as the number of concepts grows, MCD becomes more faithful. This adds a nuanced touch to the findings from the HU-MCD paper, which claimed superior faithfulness over MCD based on results from a deletion-based proxy.

\section{Conclusion}
This paper argues that we lack a clear view on how faithful current U-CBEMs really are, largely because of measures that deviate from the formal definition of faithfulness. In accordance with this definition and associated desiderata, we propose SURF as a simple, principled measure for faithfulness. Among other things, we find that \textbf{SOTA U-CBEMs do not faithfully explain the final output layer}. While SURF accurately measures U-CBEM faithfulness, we emphasize that it should be paired with an interpretability analysis to judge the U-CBEM's overall quality and value. We urge future work on concept-based explanations to adopt SURF as a standard faithfulness measure and to report SURF scores along their interpretability claims. U-CBEMs are predominantly applied on the final layer, but there is increasing interest in interpreting intermediate layers. SURF, as formulated, does not apply to this case, given the non-linear relationship between the explanation and the model's output. Extending SURF to evaluate U-CBEMs for intermediate layers is important future work.

\section{Acknowledgment}
The support of the Office of Naval Research under grant N00014-24-1-2169, and IBM-Illinois Discovery Accelerator Institute (IIDAI), and USDA National Institute of Food and Agriculture under grant AFRI 2020-67021-32799/1024178 are gratefully acknowledged. This material is based upon work supported by the National Science Foundation Graduate
Research Fellowship Program under Grant No. DGE 21-46756. Any opinions, findings,
and conclusions or recommendations expressed in this material are those of the author(s)
and do not necessarily reflect the views of the National Science Foundation. Shubham Kumar also gratefully acknowledges the support of UIUC ECE's Distinguished Research Fellowship and Promise of Excellence Fellowship.

%% file: sec/supp.tex
\maketitlesupplementary
\setcounter{page}{1}
\renewcommand{\thesection}{\Alph{section}} 
\renewcommand{\theHsection}{S\thesection}  
\setcounter{section}{0}                    

This supplementary section provides information not included in the main text due to space constraints. It includes:

\begin{enumerate}
    \item \cref{sec:appendix-sae-discussion}: Discussion of related SAE faithfulness metrics.
    \item \cref{sec:justify-surrogate-appendix}: Implementation details for the measure-over-measure study (\cref{sec:expeirment-justifying}). 
    \item \cref{sec:benchmarking-appendix}: Experimental details for the evaluation of U-CBEMs done in \cref{sec:results-benchmark}.
    \item \cref{sec:appendix-additional-benchmark-results}: Results on additional models for the experiment done in \cref{sec:results-benchmark}.
\end{enumerate}

\section{Discussion and Comparison with SAE Faithfulness Metrics}
\label{sec:appendix-sae-discussion}
There has been a significant effort in recent years to explore using SAEs for explainability, particularly on large language models \cite{bricken2023monosemanticity}. SAEs are typically trained to minimize embedding reconstruction error (e.g., mean squared error---MSE), which is the primary metric used to evaluate their faithfulness. The intuition is that if an SAE can accurately reconstruct the intermediate embedding it is explaining, then downstream operations on the reconstructed embedding should also be faithfully preserved. While reconstruction error is useful, it is a \textit{proxy} for faithfulness; there is no clear mapping from reconstruction to downstream prediction error, making it hard to interpret the significance of a given MSE error. To accommodate this, recent work \cite{gao2025scaling,kondapaneni2025representational} introduces downstream losses on model outputs to measure faithfulness. Common choices are KL divergence and cross entropy loss. However, we argue that these are uninterpretable (as acknowledged in \cite{gao2025scaling}). These metrics work well when used \textit{relatively}, which fits the referenced papers's need to compare different SAEs, but these metrics do not give an \textit{absolute} measure of faithfulness (unlike SURF), which is necessary when one wants to make an independent assessment of a method's faithfulness.

Similarly, \cite{lim2025sparse} develops a new SAE architecture (PatchSAE) and evaluates downstream classification accuracy after masking SAE latents (which is equivalent to a deletion-based approach). This metric suits their purpose, since they use SAEs to analyze model behavior, rather than to explain a prediction. \cite{fel2025archetypal} introduce a more robust SAE architecture (Archetypal SAEs) and introduce several metrics to evaluate. The closest metric to SURF is a plausibility metric, which measures the average maximum cosine similarity between the discovered CAVs and each classification vector. However, we argue that this metric does not give a holistic picture of faithfulness, since it only looks at the \textit{maximum} and it is unclear how this metric maps on to errors in the output space (e.g., how does it get transformed by a softmax). We believe that SURF is a complementary measure that can be used to help study faithfulness of SAEs applied to vision models with the intention of explaining the model's output computation. 

\section{Implementation Details for Measure-over-Measure Comparison}
\label{sec:justify-surrogate-appendix}
This section contains details for the measure-over-measure comparison conducted in \cref{sec:expeirment-justifying}.

\subsection{Experimental Setting Details}
In the \textit{Perfect} setting, we set CAVs and importances according to the weights of the final classification layer. Specifically, each class has 1 CAV ($K=1$). The CAV $\mathbf{v_{c,1}}$ for class $c$ is set to $\frac{\mathbf{f}_c}{{\|\mathbf{f}_c\|}_2}$, where $\mathbf{f}_c$ is class $c$'s classification vector. The CAV's importance $\alpha_{c,1}$ is set to ${\|\mathbf{f}_c\|}_2$.

For the \textit{random} settings, we sample each dimension of the CAVs independently from a standard normal distribution, and then make the CAV unit norm. We sample concept importances uniformly from $[0,1)$.

\subsection{Surrogate Implementation}
Recall that U-CBEMs define their own mechanism $\mathcal{P} : \mathcal{H} \to \mathbb{R}^{K}$ to project embeddings to the concept basis. For these experiments, since we are not using any U-CBEM and instead directly setting the CAVs and importances, we define $\mathcal{P}(\mathbf{h}_{j}; V_i)_k = \mathbf{h}_{j}^T \mathbf{v}_{i,k}$, where $i$ denotes the class, $k$ denotes the CAV, and $j$ selects the spatial embedding. Recall in this case that $K=1$ and there is only one embedding (obtained from global pooling).

\subsubsection{SURF Surrogate (Ours)}
Following \cref{sec:method-surrogate} and \cref{eq:sf-add}, we set the parameters of the surrogate with the obtained concepts and importances. This surrogate has no learnable parameters and is exactly equal to the original linear layer (up to the bias term). Note that the embeddings are linearly projected onto the CAVs to find the concept representation. After the surrogate reconstructs the output, we add the constant bias term before evaluation.

\subsubsection{ICE Surrogate}
ICE's surrogate takes the concept representation and reconstructs the embedding, as specified by the U-CBEM. It then passes through the final classification layer to obtain outputs. Let $\mathbf{p_i} \in \mathbb{R}^K$ denote the concept representation for class $i$ obtained from embedding $\mathbf{h}$. The surrogate is:
\begin{equation}
    \begin{aligned}
            y_i &= \mathbf{f}_i^T \psi_{ICE}(\mathbf{p_i}) + b_i
    \end{aligned}
\end{equation}
where $\psi_{ICE}(.)$ denotes the reconstruction, as specified by the U-CBEM. Since we are not making use of any U-CBEM in the measure-over-measure comparison, we use linear operations to reconstruct the embedding from the concept representation. That is, $\psi(\mathbf{p_i}) = \sum_{k=1}^K p_{i,k}\mathbf{v}_{i,k}$.

\subsubsection{C-SHAP Surrogate}
C-SHAP's surrogate defines a two-layer MLP $\psi_{\text{C-SHAP}}(.)$ to reconstruct the embedding, and then passes through the final classification layer to obtain outputs. The surrogate is:
\begin{equation}
    \begin{aligned}
            y_i &= \mathbf{f}_i^T \psi_{\text{C-SHAP}}(\mathbf{p_i}) + b_i
    \end{aligned}
\end{equation}
Since C-SHAP's surrogate contains learnable parameters. Following \cite{ConceptSHAP}, the MLP's hidden layer has a dimensionality of 500, and the surrogate is trained using Cross Entropy Loss between the surrogate's predicted class probabilities and the original model's class probabilities. We train the surrogate using the Adam optimizer \cite{kingma2014adam} and a starting learning rate of $0.1$. The surrogate is trained for at most 100 epochs, during which the learning rate is decayed by a factor of 0.5 when the training loss plateaus. We stop training when the learning rate is reduced below $10^{-7}$.

\subsection{FLOPs Computation}
We treat multiply and add as 2 FLOPs. Let $K$ be the number of concepts, $C$ be the number of output classes, and $D$ by the dimensionality of the embedding space.

\textbf{SURF Surrogate:} Given that the surrogate is a linear layer from concept representations to the output space (though it uses different concept representations for each class), the number of FLOPs required is $2KC$.

\textbf{ICE Surrogate:} This surrogate reconstructs the original embedding (by a linear projection) from the concept representation, and then uses the original model (which is also a linear projection) to map to the output space. Thus, the number of FLOPs required is $CD(2K+1)$.

\textbf{C-SHAP Surrogate:} This surrogate reconstructs the original embedding (with a two-layer MLP) from the concept representation, and then uses the original model (which is a linear projection) to map to the output space. Thus, the number of FLOPs required is $2C(H_1K + H_1D + D)$, where $H_1$ is the dimensionality of the MLP's hidden layer.

\section{Additional Details for Benchmarking U-CBEM Faithfulness}
\label{sec:benchmarking-appendix}
\cref{sec:results-benchmark} evaluates prior U-CBEMs, finding that they produce unfaithful explanations. This section gives experimental and implementation details for this evaluation.

\subsection{Model Details}
\textbf{Object Classification:} Our evaluation uses a ResNet-50, available from the \texttt{timm} library. We use the ImageNet pre-trained weights and initialize the final classification layer with random weights. The model is finetuned (\texttt{conv4} and all subsequent layers) on the Caltech-101 training set, achieving a test accuracy of approximately 95.61\%. The model is trained without data augmentation and using cross entropy loss with label smoothing. 

\textbf{Multi-Attribute Prediction:} We finetune an ImageNet pre-trained MobileNetV2 on the CelebA dataset for the task of binary attribute prediction. We initialize the final prediction linear layer with random weights, and finetune all layers of the model. We achieve an attribute prediction accuracy of 91.55\%. The model is trained with data augmentation and binary cross-entropy loss.

\textbf{Age Regression:} We finetune an ImageNet pre-trained Vision Transformer (ViT) on the UTK-Face dataset for the task of age regression. We choose the ViT-B/16 variant for our experiment. We initialize the final prediction linear layer with random weights, and finetune all weights above (not including) the 9th layer. The model achieves a mean average error of 5.14 on the test set. The model is trained with data augmentation and using L1 Loss.

\subsection{Dataset Details}
\label{sec:benchmarking-appendix-eval-details}
We learn each class's concepts and importances using images of the class from the training set. Importantly, the label of each image is determined using the original model's prediction (instead of the ground truth label). We do this because we are interested in explaining why \textit{the model} predicts a specific label, regardless of whether the prediction is correct or not. After learning the U-CBEMs, we evaluate them on a test set. 

\textbf{Object Classification:} We split Caltech-101's dataset into train and test sets using 80/20 splits. While splitting the data, we maintain the same class distribution present in the original dataset in both the training and testing splits. To handle Caltech-101's class imbalance, we assemble a balanced test set; this ensures that we evaluate for faithfulness equally across all classes. 

\textbf{Multi-Attribute Prediction:} We use CelebA's pre-defined train and test sets. We do not make any modifications to the test set. Since the training set is too large for many U-CBEMs, we take a random subset of training data to use for finding U-CBEM parameters. The random subset is equivalent in size to the test set.

\textbf{Age Regression:} We split the dataset into train and test sets using 80/20 splits. While splitting the data, we discretize the age distribution into bins of 10-years, and maintain this distribution across both the training and testing splits. As UTK-Face does not have any classes, all U-CBEMs find CAVs on all embeddings.

\subsection{U-CBEM Details}
This section specifies implementation details for each U-CBEM that was benchmarked.

\subsection{CDISCO}
\href{https://github.com/maragraziani/concept_discovery_svd}{The open-source implementation} provided by the authors was used to evaluate this approach. CDISCO uses SVD to discover CAVs. Then, a gradient-based approach is used to calculate the importance of each CAV. The code used to compute the gradients was not provided in the repository, so we re-implemented it. Embeddings are linearly projected to the concept space.

\textbf{Object Classification:} Due to a lack of clarity, we had to make one implementation choice. Specifically, when finding the importance (using a gradient calculation), it was unclear if we should use images from all classes or only images of the CAV's class. We opted for the latter, based on the code and the example provided by the authors. 

\textbf{Multi-Attribute Prediction:} CDISCO was extended to this task by reformulating the task as independent binary classification tasks. Thus, CAVs and concept importances were learned for each task independently

\textbf{Age Regression:} The open-source implementation only considered the case of multi-class classification. According to the paper and the code used for multi-class classification, we re-implemented CDISCO to work in the case of single-output regression.

\subsection{ICE}
\href{https://github.com/zhangrh93/InvertibleCE/tree/main}{The open-source implementation} provided by the authors was used to evaluate this approach. ICE uses NMF to find CAVs. Due to the potentially large number of embeddings, we use the batched-version of NMF. Concept importance is found using the TCAV method proposed in \cite{kimTCAV}. Embeddings are projected to the concept space using sci-kit learn's transform method. ICE was extended to multi-attribute prediction by reformulating the task as independent binary classification tasks. Thus, CAVs and concept importances were learned for each task independently.

\subsection{CRAFT}
The open-source implementation provided in the Xplique toolbox \cite{fel2022xplique} (by the same authors) was used to evaluate this approach. CRAFT uses NMF to find CAVs and introduces a Sobol-based sensitivity approach to find concept importance. Embeddings are projected to the concept space using the toolbox's built-in transform method.

\subsection{ConceptSHAP}
ConceptSHAP differs from previous U-CBEMs because it discovers CAVs globally (i.e., class-agnostic). Thus, we need a way to associate discovered CAVs to the concepts. To do so, we discover a (larger than $K$) set of CAVs and assign them a per-class concept importance score (through Shapley Values). Then, the top $K$ CAVs by importance magnitude are selected to be the CAVs for a given class. For all experiments, we initially discover 100 CAVs. This number is arbitrary, but note that the Shapley Value computation scales exponentially with the number of initially discovered CAVs, which motivated our choice. With more CAVs, we expect ConceptSHAP to obtain higher faithfulness at the expense of a higher computational cost. Embeddings are linearly projected to the concept space. For age regression, we only discover $K$ concepts (since there are no classes). For multi-attribute prediction, we repeat the ConceptSHAP procedure for each attribute, considering each attribute prediction as a binary classification task; we only discover $K$ concepts, since each task is a binary classification.

The authors provide an open-source implementation in TensorFlow, so we instead use a \href{https://github.com/arnav-gudibande/conceptSHAP/tree/master}{third-party re-implementation} in PyTorch. Several modifications are made to this implementation:

\begin{enumerate}
    \item Added in a learnable two-layer MLP to map back to the representation space (as done in the original paper)
    \item Added more efficient concept importance scoring via Shapley Values. This is implemented using KernelSHAP (\cite{shapley}) and is class-specific.
    \item Fixed the implementation of the completeness score
\end{enumerate}

\begin{table*}[tb]
  \caption{\textbf{Additional Benchmark Results.} We apply SURF to evaluate explanations from prior U-CBEMs on the \textit{Object Classification} task from \cref{sec:results-benchmark} on a VGG11 and InceptionV3. Along with the $\text{SURF}_{\text{MAE}}$ and $\text{SURF}_{\text{EMD}}$, we report other metrics to serve as a comparison. Our trends hold: prior U-CBEMs are not faithful, as indicated by large errors in the logit and probability space.}
  \label{tab:appendix_benchmark}
  \centering
  \small
  \begingroup  
  \footnotesize

  \begin{subtable}[t]{0.6\textwidth}
    \centering
    \subcaption{Object classification \\ (VGG11)}
    \label{tab:appendix_vgg}
    \begin{tabular}{lccccc}
      \toprule
      \textbf{U-CBEM} & $\text{\textbf{SURF}}_{\text{MAE}}$ ($\downarrow$) & $\text{\textbf{SURF}}_{\text{EMD}}$ ($\downarrow$) &
        \textbf{Top-1 (\%)} ($\uparrow$) & \textbf{Rank Corr} ($\uparrow$) \\
      \midrule      
      CDISCO & 2.89 & 0.936 & 0.4 & -0.01 \\
      ICE & 3.58 & \underline{0.616} & \textbf{93.7} & \underline{0.498} \\
      CRAFT & 2.50 & 0.844 & 51.0 & 0.214 \\
      C-SHAP & 3.28 & 0.884 & 1.1 & -0.00 \\
      MCD & \underline{1.95} & 0.738 & \underline{83.3} & 0.306 \\
      HU-MCD & -- & -- & -- & -- \\
      SAE & \textbf{0.47} & \textbf{0.567} & 62.4 & \textbf{0.504} \\
      \bottomrule
    \end{tabular}
  \end{subtable}%
  \hfill
  \begin{subtable}[t]{0.40\textwidth}
    \centering
    \captionsetup{justification=centering} 
    \subcaption{Object classification \\ (InceptionV3)}
    \label{tab:appendix_inception}
    \begin{tabular}{ccccc}
      \toprule
      $\text{\textbf{SURF}}_{\text{MAE}}$ & $\text{\textbf{SURF}}_{\text{EMD}}$ &
        \textbf{Top-1 (\%)} & \textbf{Rank Corr} \\
      \midrule      
      4.08 & 0.847 & 1.52 & 0.012 \\
      3.34 & 0.595 & 50.3 & -0.05 \\
      4.26 & 0.652 & 8.41 & 0.131 \\
      4.31 & 0.754 & 9.70 & -0.08 \\
      \underline{2.76} & \underline{0.327} & \textbf{89.1} & \underline{0.385} \\
      -- & -- & -- & -- \\
      \textbf{0.35} & \textbf{0.226} & \underline{72.0} & \textbf{0.770} \\
      \bottomrule
    \end{tabular}
  \end{subtable}%
  \hfill

\endgroup
\end{table*}

\subsection{MCD}
Instead of representing concepts as single CAVs, MCD represents concepts as multi-dimensional subspaces. In \cite{vielhaben2023multidimensional}, any concept $l$ for class $i$ is characterized by a representative basis $C^{i,l} = \{\mathbf{v}_j^{i,l}|j=1...d_l\}$, where $d_l$ denotes the dimensionality of the subspace. To ensure the union of all concepts span the entire feature space, they define the orthogonal complement concept $C^{i,\perp} = \text{span}(C^{i,1},...,C^{i,K})^\perp$, where $K$ is the number of concepts originally discovered, letting $C^{i,K+1} \equiv C^{i,\perp}$. Then, any embedding $\mathbf{h}$ can be uniquely decomposed onto the concept basis as:
\begin{equation}
    \begin{aligned}
        \mathbf{h} = \sum_{l=1}^{n_c+1} \sum_{j=1}^{d^{\,l}} p_j^{i,l}\, \mathbf{v}_j^{i,l}
    \end{aligned}
\end{equation}
Thus, we stack all $p_j^{i,l}$ into the concept representation $\mathbf{p_i}$ for class $i$. Since $C^{i,\perp}$ is not interpretable, we set elements of $\mathbf{p_i}$ corresponding to $C^{i,\perp}$ to 0; thus, the reconstruction of $\mathbf{h}$ is done using interpretable concepts presented to the user during an explanation. We set concept importance $\alpha_j^{i,l} = \mathbf{f}_i^T \mathbf{v}_j^{i,l}$, which directly relates to the local and global concept relevance defined in MCD and allows for faithful reconstruction of the output using our surrogate. 

Since MCD is computationally expensive when trying to discover concepts on a large number of embeddings, we randomly sample 10,000 embeddings if there are greater than these many embeddings extracted for a given class. We use the \href{https://github.com/jvielhaben/MCD-XAI}{open-source implementation} provided by the authors.

\subsection{HU-MCD}
HU-MCD directly extends upon MCD, adopting the same framework of concept subspace, orthogonal complement, and unique decomposition onto the concept basis. The only material difference is in the concept discovery stage, where concepts are discovered on representations arising from image segments, instead of from the feature map. Therefore, we make the same adaptations as MCD for HU-MCD. We use the \href{https://github.com/grobruegge/hu-mcd}{open-source implementation} provided by the authors.

\subsection{SAE}
We learn a Top-K SAE \cite{gao2025scaling} on the training embeddings (before average pooling) using the Overcomplete library. SAEs learn class-agnostic concepts (in contrast to U-CBEMs that learn CAVs for a specific class output); to enable a fair comparison, our SAE learns the same number of total CAVs as the other U-CBEMs for each setting. As an example, our SAE for the \textit{Object Classification} task learns 505 CAVs (5 per class output, 101 total classes). At reconstruction time, we set the $K$ in the Top-K SAE to 5, directly mirroring the other U-CBEMs.

\section{Additional Benchmark Results}
\label{sec:appendix-additional-benchmark-results}
We repeat our experiment on the \textit{Object Classification} task from \cref{sec:results-benchmark} on a VGG11 and InceptionV3 to confirm trends hold. We use the ImageNet pre-trained weights, and finetune them for the \textit{Object Classification} task. Results are reported in \cref{tab:appendix_benchmark}. For both models, SAE is the most faithful. For VGG11, we find that all U-CBEMs tend to exhibit large EMD errors (larger than 0.5). Though SAE is able to reduce the logit error significantly, the metrics indicate that there are still issues when trying to reproduce predictions in the probability space. While ICE can accurately reconstruct the Top-1 prediction, it has large logit errors. For InceptionV3, we find ICE is not faithful, and instead MCD is second-best (after SAE). Compared to the ResNet studied in \cref{sec:results-benchmark}, we believe that the models tested here have more nuance in their predicted probability space, which may lead to larger EMD errors, even if the logit error seems small.